# Diffusion-enabled Optimal Transport distances for graph matching

Iman Seyedi[1[0000-0003-1875-7565]] and Francesco Archetti[1[0000-0003-1131-3830]]

[1] Department of Computer Science Systems and Communication, University of Milano-Bicocca, 20126 Milan, Italy

seyediman.seyedi@unimib.it francesco.archetti@unimib.it

**Abstract.** This paper introduces Diffusion Semi-Relaxed Fused Gromov-Wasserstein (DsrFGW), a novel method for graph comparison that unifies node features and structural connectivity through optimal transport. While traditional Gromov-Wasserstein and semi-relaxed variants (srGW, srFGW) capture graph structure, they often struggle with sparse, noisy, or partially observed graphs. Inspired by Graph Diffusion Distance—which posits graphs are similar if they enable similar information transmission patterns—DsrFGW incorporates diffusion processes allowing information propagation across nodes, capturing local and global structural patterns while reducing sensitivity to noise or missing edges. An extensive evaluation on 36 synthetic pairwise graph matching tasks (easy, medium, hard) demonstrates consistent superiority over srFGW, achieving accuracy improvements of 0-20 percentage points and dramatic Adjusted Rand Index (ARI) gains: in medium-difficulty scenarios, srFGW often achieves negative ARI (worse than random) while DsrFGW offers better performance in terms of both internal and external clustering quality measures (i.e., Adjusted Rank Index and Accuracy with respect to the true underlying clusters, respectively). Even under severe noise, DsrFGW improves clustering quality in 92% of the synthetic tasks with optimal diffusion scales adapting to problem difficulty, establishing DsrFGW as a robust framework for graph comparison under structural uncertainty.

## 1 Introduction

Graph-structured data arises naturally in many domains, including neuroscience, social networks, and transportation systems. Nodes carry descriptive features while edges encode structural relationships, making graph comparison central to machine learning and network analysis. Traditional supervised alignment methods, relying on predefined correspondences, often fail with noisy, sparsely connected, or partially observed networks. Optimal Transport (OT) provides a principled framework for comparing probability measures, while its extension—Gromov-Wasserstein (GW) distances—enables graph comparison based on structure alone, without requiring direct node-to-node correspondences. Semi-relaxed variants (srGW, srFGW) further improve computational efficiency and allow incorporation of node features through fused formulations [1], successfully applied to unsupervised alignment of brain connectivity networks [2] and other node-attributed graphs.

Graph Diffusion Distance (GDD) measures graph similarity by comparing information propagation across networks [3], [4]. Diffusion distances capture local and global structural patterns, are robust to edge deletions, and complement OT-based methods by emphasizing graph dynamics rather than static connectivity alone. The core intuition is that node features should incorporate information from their neighborhood through the graph structure. Figure 1 illustrates this process: starting from localized feature values ($\tau = 0$), the heat kernel diffusion operator gradually propagates information along graph edges. At small diffusion parameter ($\tau = 0.05$), features spread to immediate neighbors, while larger diffusion ($\tau = 0.5$) induces global smoothing that captures multi-hop structural relationships, effectively enriching feature representation by encoding graph topology directly into node attributes.

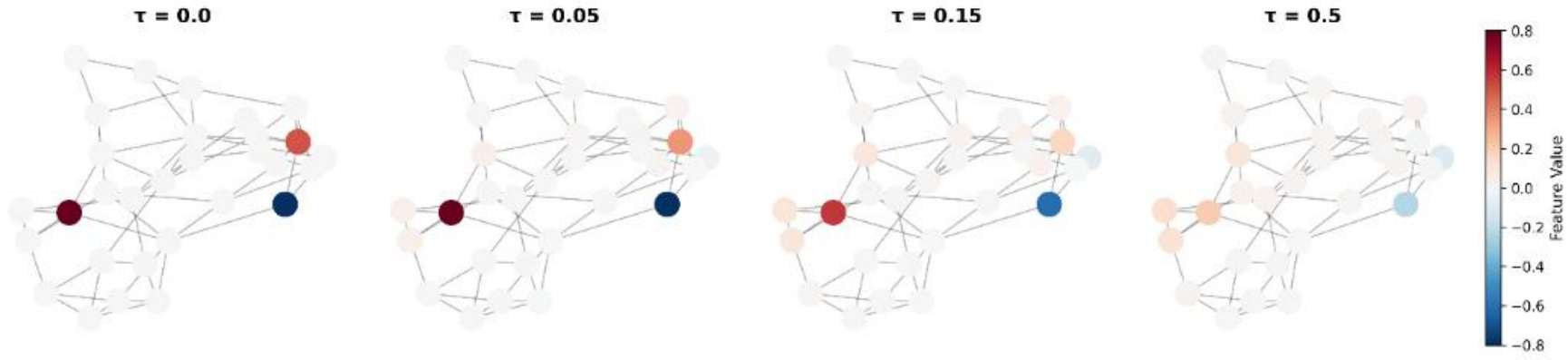

**Fig. 1.** Heat kernel diffusion on graphs. Initially ($\tau = 0$), three nodes have distinct feature values (red +1.0, blue -1.0, light red +0.5), others are neutral. As $\tau$ increases, features spread through edges: $\tau = 0.05$ affects immediate neighbors, $\tau = 0.15$ extends to multi-hop neighborhoods, $\tau = 0.5$ produces global smoothing across the graph.

Recent work on Wasserstein diffusion on multidimensional spaces [5], [6] provides theoretical foundation for integrating diffusion processes into OT-based frameworks. We introduce Diffusion Semi-Relaxed Fused Gromov-Wasserstein (DsrFGW), integrating diffusion-based transformations of node features with semi-relaxed fused GW transport. By diffusing node attributes before computing the transport plan, DsrFGW leverages both connectivity and feature information, improving alignment accuracy particularly for sparse or noisy networks. The method bridges ideas from computational optimal transport [7] and attributed graph clustering [8], extending classical results on OT-based learning of probability measures [9]. Our main contributions are: (1) Methodological innovation: DsrFGW unifies graph diffusion with semi-relaxed fused Gromov-Wasserstein optimal transport for robust alignment under noise and sparsity; (2) Empirical validation: Through systematic experiments on 36 synthetic configurations, we demonstrate substantial improvements over srFGW baselines, with accuracy gains up to 48% and distance reductions up to 61%, particularly in noisy regimes; (3) Practical guidelines: We characterize sensitivity of DsrFGW to key parameters (diffusion $\tau$, structure weight $\alpha$) across varying graph sizes and difficulty levels.

The paper is structured as follows: Section 2 reviews related work; Section 3 introduces mathematical background; Section 4 presents the DsrFGW algorithm; Section 5 reports numerical experiments; Section 6 draws conclusions and outlines future perspectives.

## 2 Related Works

Graph comparison and alignment are central challenges across neuroscience, transportation, and social networks. Optimal Transport (OT) provides a natural framework for comparing distributions, with GW distance enabling graph comparison by focusing on relational structure between nodes rather than requiring explicit correspondences. GW has been successfully applied to clustering, coarsening, and embedding of graphs [11], [12], [13]. While GW captures structural information, it ignores node attributes. Fused Gromov-Wasserstein (FGW) addresses this by combining GW over graph structure with Wasserstein distance over node features, making it suitable for node-attributed graphs such as traffic networks or functional brain connectivity networks [8], [14]. The srFGW improves scalability and handle graphs of different sizes by relaxing target graph weights during alignment, reducing computational constraints while enabling efficient optimization [1].
Complementing OT-based approaches, Graph Diffusion Distance (GDD) measures similarity by modeling signal spread via the heat equation over the graph Laplacian [3], [4]. Diffusion distances are robust to structural perturbations and provide insight into local and global connectivity patterns. Recent works have combined diffusion and OT concepts, proposing Diffusion Wasserstein distances that embed node features and diffusion patterns into a unified distance [3], [5], [6]. Theoretical foundations for combining diffusion processes with optimal transport have been established through several contributions: [15] pioneered Wasserstein diffusion processes on probability measure spaces; [16] developed alternative constructions through coalescing particle systems; [17] generalized the framework to p-Wasserstein spaces over Banach spaces; [6] established reversible diffusion processes on multidimensional Wasserstein spaces; and [18] analyzed convergence properties of diffusion-based models in Wasserstein distances.
Our work introduces DsrFGW, which extends srFGW by incorporating node-level diffusion into the transport plan, capturing global graph structure, node attributes, and information propagation patterns. While theoretical works establish diffusion processes on Wasserstein spaces themselves, our approach applies diffusion directly to node features within individual graphs before computing the transport plan. DsrFGW unifies structure-based alignment (GW), attribute-aware alignment (FGW, srFGW), and dynamics-based similarity (diffusion distances), situating itself within recent advances in graph clustering, alignment, and distributional reduction [8], [9], [19].

## 3 Mathematical Background

This section introduces the foundational concepts underlying our approach, including Optimal Transport (OT), metric measure spaces, and semi-relaxed Gromov-Wasserstein distances for node-attributed graphs. These tools form the basis for our proposed DsrFGW divergence.

### 3.1 Optimal Transport

Optimal Transport (OT) provides a mathematical framework for comparing probability measures while accounting for the geometry of their underlying spaces. The problem originates in the work of [20], who formulated transport as finding a deterministic map minimizing total displacement cost between metric spaces $(X, d)$ and $(Y, d')$. Kantorovich [21] introduced a relaxation allowing mass splitting through transport plans $\gamma \in \Pi(\mu, \nu)$, yielding the discrete formulation $\min_{T \in \mathbb{R}_+^{n\times m}} \sum_{i=1}^{n} \sum_{j=1}^{m} c(x_i, y_j)\, T_{ij}$ subject to marginal constraints $T 1_m = h,\ T^\top 1_n = h'$, which replaces permutations by soft couplings as a convex linear program that always admits solutions [7]. This framework underlies Wasserstein distances $W_q^q(\mu, \nu) = \min_{\pi \in \Pi(\mu,\nu)} \int_{\mathcal{X}\times\mathcal{Y}} d(x, y)^q \, d\pi(x, y)$, which define meaningful metrics between probability measures respecting the geometry of the sample space and enable interpolation, barycenters, and gradient flows [7], [22], [23], [24], [25], [26], [28]. The 2-Wasserstein distance admits an equivalent Monge formulation with closed-form solutions for Gaussian distributions via the Bures metric, while general distributions require approximate solutions such as entropic regularization and Sinkhorn-based algorithms.

### 3.2 Gromov–Wasserstein and fused Gromov–Wasserstein distance

In the discrete graph setting, mm-spaces provide a natural representation of weighted graphs with node distributions [28]. A graph $G = (V, E, w)$ is encoded as a metric measure space $(V, C, p)$ where V is the node set, $C \in \mathbb{R}^{n\times n}$ is a distance matrix (e.g., shortest-path, resistance, or diffusion distance), and $p \in \Sigma_n$ is a probability vector over nodes (e.g., uniform or degree-based). The Wasserstein distance between two mm-spaces $(V, C, p)$ and $(V', C', p')$ quantifies the "effort" to transform one graph's node distribution into the other's, accounting for both geometry (via $C, C'$) and node masses (via $p, p'$). However, standard Wasserstein distance requires graphs to share a common ambient space for comparing node positions. The GW distance overcomes this limitation by comparing the internal metric structures of graphs rather than absolute positions [5], [11], [29].

Let $G = (C, h)$ and $G' = (C', h')$ be two graphs represented by pairwise dissimilarity matrices $C \in \mathbb{R}^{n\times n}$, $C' \in \mathbb{R}^{m\times m}$, and node distributions $h \in \Sigma_n$, $h' \in \Sigma_m$. The p-Gromov–Wasserstein distance is defined as

$$GW_q^q(C, h, C', h') = \min_{T \in U(h,h')} \sum_{i,j,k,l} \mid C_{ij} - C'_{kl} \mid^q T_{ik} T_{jl} \quad (1)$$

where $U(h,h') = \{T \in \mathbb{R}_+^{n\times m} \mid T1_m = h,\ T^\top 1_n = h'\}$ ensures the coupling T satisfies both marginal constraints. GW captures structural similarity but ignores node attributes. To jointly consider structure and features, FGW distances interpolate between GW and Wasserstein distances on node features [14]. Let $X \in \mathbb{R}^{n\times r}$, $Y \in \mathbb{R}^{m\times r}$ be node feature matrices and define the feature cost:

$$M_{ij}^p = \| x_i - y_j \|_p^p \tag{2}$$

The FGW distance is

$$FGW_{q,\alpha}^q\,(C,F,h,C',F',h') = \min_{T\in U(h,h')} \langle (1-\alpha)M^q(F,F') + \alpha L_q(C,C') \otimes T, T\rangle \tag{3}$$

where $\alpha \in [0,1]$ balances structure versus features, and $L_q(C,C') \otimes T$ denotes the standard GW structural term [14]. Figure 2 from [14] illustrates the key idea behind the FGW distance. It visually represents how FGW aligns nodes between two graphs by considering both feature similarity (e.g., node attributes or embeddings) and structural similarity (relationships between nodes captured by the adjacency or structure matrices). Figure 2 from [14] illustrates the key idea behind the FGW distance.

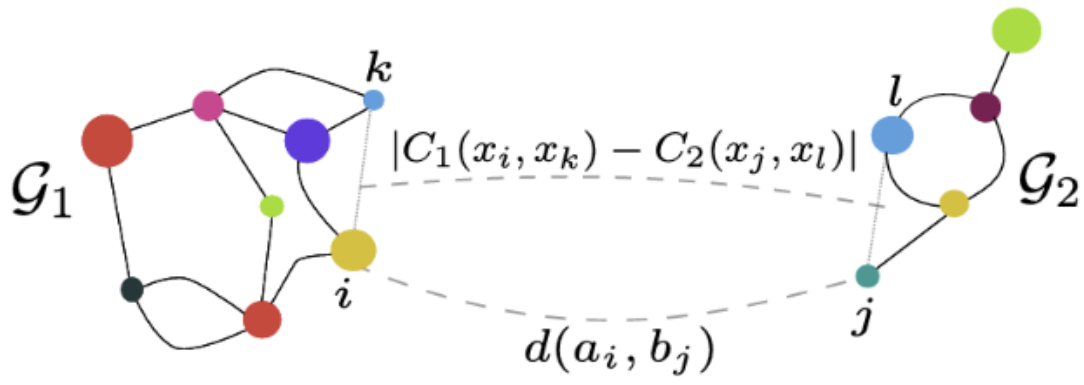

**Fig. 2.** Fused Gromov-Wasserstein for attributed graphs (from [14]). Node colors represent features, node sizes indicate probability masses. FGW jointly optimizes feature similarity d(a_i,b_j) and structural consistency $|C_1(x_i,x_k) - C_2(x_j,x_l)|$, with parameter α balancing these dual objectives.

## 3.3 Semi-Relaxed Gromov-Wasserstein (srGW) Divergence

In many applications, the target node distribution h' is unknown or ill-defined. Semi-Relaxed GW (srGW) [1] relaxes the second marginal $h'$, allowing automatic determination of relevant target nodes and discarding irrelevant parts of the target graph. The admissible set becomes $U(h,m) = \{T \in R_+^{n\times m} \mid T1_m = h\}$, and the srGW discrepancy is

$$srGW_q^q(C,h,C') = \min_{T\in U(h,m)} \sum_{i,j,k,l} | C_{ij} - C'_{kl} |^q\, T_{ik}T_{jl} \tag{4}$$

with optimal target weights recovered as $h'^{*} = T^{*\top} 1_n$. This approach naturally selects meaningful subgraphs and produces sparse solutions. Figure 3 [1] illustrates this: when the source has two clusters and the target has three, classical GW forces uniform distribution across all clusters (high cost), while srGW optimizes the target distribution to ignore one cluster and focus on the two best-matching clusters, cleanly aligning source clusters with a target subgraph at significantly reduced cost.

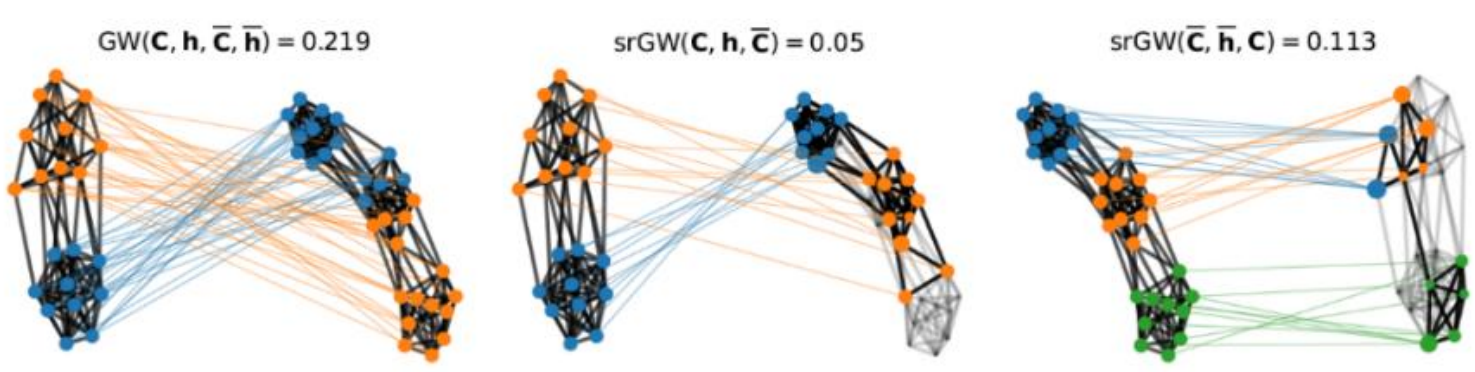

**Fig. 3.** Classical GW vs. srGW (from [1]). Source (two clusters) matched to target (three clusters). Left: Classical GW forces uniform distribution (cost = 0.219). Middle: srGW relaxes target marginals (cost = 0.05). Right: srGW relaxes source marginals (cost = 0.113).

### 3.4 Semi-relaxed fused Gromov–Wasserstein divergence

Extending the srgw to attributed graphs yields the semi-relaxed fused Gromov–Wasserstein (srFGW) problem. SrFGW extends srGW to node-attributed graphs by combining structural and feature information. Let $G = (C, X, h), G' = (C', Y)$ where no target distribution is specified. The srFGW distance for $p \geq 1$ and $\alpha \in [0,1]$ is defined as

$$srFGW_{q,\alpha}^{q} = \min_{T \in U(h,m)} \{(1-\alpha) \sum_{ij} \| F_i - F_j' \|_q^q \, Tij + \alpha \sum_{ijkl} | C_{ij} - C_{kl}' |^q \, T_{ik} T_{jl}\} \tag{5}$$

Where $U(h, m) = \{ T \in \mathbb{R}_+^{n \times m} \mid T 1_m = h \}$ and the optimal target measure is recovered as $h'^{*} = T^{*\top} 1_n$. it reduces to FGW when target marginal $h'$ is fixed, focuses solely on features (Wasserstein) when α=0 or structure (srGW) when α=1, automatically selects important nodes producing sparse solutions, and is computationally cheaper than FGW as only one marginal is constrained [1], [3]. By combining semi-relaxed transport with feature-structure fusion, srFGW provides a flexible and interpretable distance suitable for graph matching, partitioning, and dictionary learning. This forms the theoretical basis for our proposed DsrFGW, which further incorporates diffusion-based node embeddings to capture information propagation across the graph [3], [5].

### 3.5 Graph diffusion and heat kernel representations

Let $G_s = (V_s, C_s, X_s, p_s)$ be a weighted attributed graph with n nodes, where $C_s \in \mathbb{R}^{n \times n}$ encodes structural dissimilarities, $X_s \in \mathbb{R}^{n \times d}$ denotes node attributes, and $p_s \in \Delta_n$ is a probability measure on nodes. The structural dissimilarity matrix C can be derived from geometric distances (shortest-path distance $C_{ij}^{SP} = \min_{paths\, \pi: i \to j} | \pi |$ computed via Floyd-Warshall or Dijkstra, or resistance distance $C_{ij}^{Res} = (L^{+})_{ii} + (L^{+})_{jj} - 2(L^{+})_{ij}$ based on the graph Laplacian pseudoinverse) or diffusion-based distances that measure similarity through connectivity patterns revealed by random walks or heat diffusion processes [30], [31], [4]. Following [4] and [3], we represent node attributes through heat kernel diffusion: for diffusion scales $\tau^s, \tau^t \geq 0$, the diffused feature representations are

$$\tilde{X} = exp(-\tau^s L^s)\, X \in \mathbb{R}^{n \times d} \text{ , } \tilde{Y} = exp\,(-\tau^t L^t) Y \in \mathbb{R}^{m \times d} \quad (6)$$

where $e^{-\tau L}$ is the heat kernel operator modeling information propagation with τ controlling spatial extent (typical values range from $\tau \sim 1/\bar{d}$ for local diffusion to $\tau \sim D$ for global diffusion). Given diffused features, we define a diffusion-dependent feature cost matrix

$$\widetilde{M}_{ij}^{p}(\tau^s, \tau^t) = \| \tilde{X}_s^{(i)} - \tilde{X}_t^{(j)} \|_p^p \quad (7)$$

where $\tilde{X}_s^{(i)} \in \mathbb{R}^d$ and $\tilde{X}_t^{(j)} \in \mathbb{R}^d$ denote the i-th and j-th rows. This matrix incorporates neighborhood context: nodes in similar structural positions have similar diffused features even if original features differ, with heat kernel stability ensuring robustness to perturbations [3]. This diffusion-enhanced representation forms the foundation for DsrFGW.

## 4. Diffusion Semi-Relaxed Fused Gromov–Wasserstein (DsrFGW)

This section introduces the DsrFGW framework, combining diffusion-based graph representations, semi-relaxed optimal transport, and fused Gromov–Wasserstein discrepancies for attributed graphs. The key idea is to replace raw node attributes with their diffused counterparts, thereby combining robustness to noise with flexible structural alignment. Given diffusion $(\tau^s, \tau^t)$, the DsrFGW discrepancy is defined as

$$DsrFGW_{p,\alpha}^{p}(\,C, X, h, C', Y \mid \tau^s, \tau^t\,) = \min_{T \in U(h,m)} \left\{ (1-\alpha)\langle \widetilde{M}^p, T \rangle + \alpha \textstyle\sum_{ijkl} \mid C_{ij} - C_{kl'} \mid^p T_{ik} T_{jl} \right\} \quad (8)$$

This couples diffusion-based feature alignment with semi-relaxed structural matching. The optimization remains a non-convex quadratic program with a single marginal constraint, reducing computational burden and improving scalability. The optimal transport plan $T^*$ induces a target distribution $h'^*(\tau^s, \tau^t) = T^{*\top} 1_n$ whose sparsity reveals which target nodes are most relevant for matching the source graph at chosen diffusion scales.

Numerical resolution follows standard optimization strategies for GW-type objectives, including conditional gradient or proximal schemes [7], with the diffusion step efficiently approximated using truncated spectral decompositions or Chebyshev polynomial expansions [3].

## Numerical Experiments

### 5.1 Experimental Setup

We evaluate the proposed DsrFGW method on synthetic pairwise graph matching tasks with systematically varying difficulty levels and scales. The experimental design allows us to assess the impact of graph diffusion on matching performance under different structural and feature noise conditions.

**Graph Generation.** Graphs are generated using a stochastic block model with two equal-sized clusters, with difficulty controlled through intra-cluster probability ($p_{in}$), inter-cluster probability ($p_{out}$), and feature noise standard deviation ($\sigma$): Easy ($p_{in} = 0.4$, $p_{out} = 0.05, \sigma = 1.0$), Medium ($p_{in} = 0.3, p_{out} = 0.08, \sigma = 2.0$), and Hard ($p_{in} = 0.2, p_{out} = 0.12, \sigma = 3.0$). As difficulty increases, cluster boundaries become less distinct both structurally (smaller $p_{in}/p_{out}$ ratio) and in feature space (larger σ), tested across three graph sizes $n \in \{50,100,200\}$ yielding 9 configurations. For each graph pair, we perform a grid search over the diffusion parameter $\tau \in \{0.0,0.01,0.05,0.1,0.2,0.5\}$ and structure-feature trade-off $\alpha \in \{0.1,0.3,0.5,0.7\}$, with baseline srFGW corresponding to $\tau = 0$. Node features are initialized as $X_i = 2L_i - 1 + \sigma \cdot \epsilon_i$, where $L_i \in \{0,1\}$ is the cluster label and $\epsilon_i \sim N(0,1)$, then normalized to zero mean and unit variance, with graph structure encoded using shortest-path distances via the Floyd-Warshall algorithm. Figure 4 illustrates representative source-target graph pairs for $n = 50$ across all difficulty levels, clearly showing degradation of cluster cohesion from dense intra-cluster connections in easy graphs to substantial mixing between clusters in hard graphs.

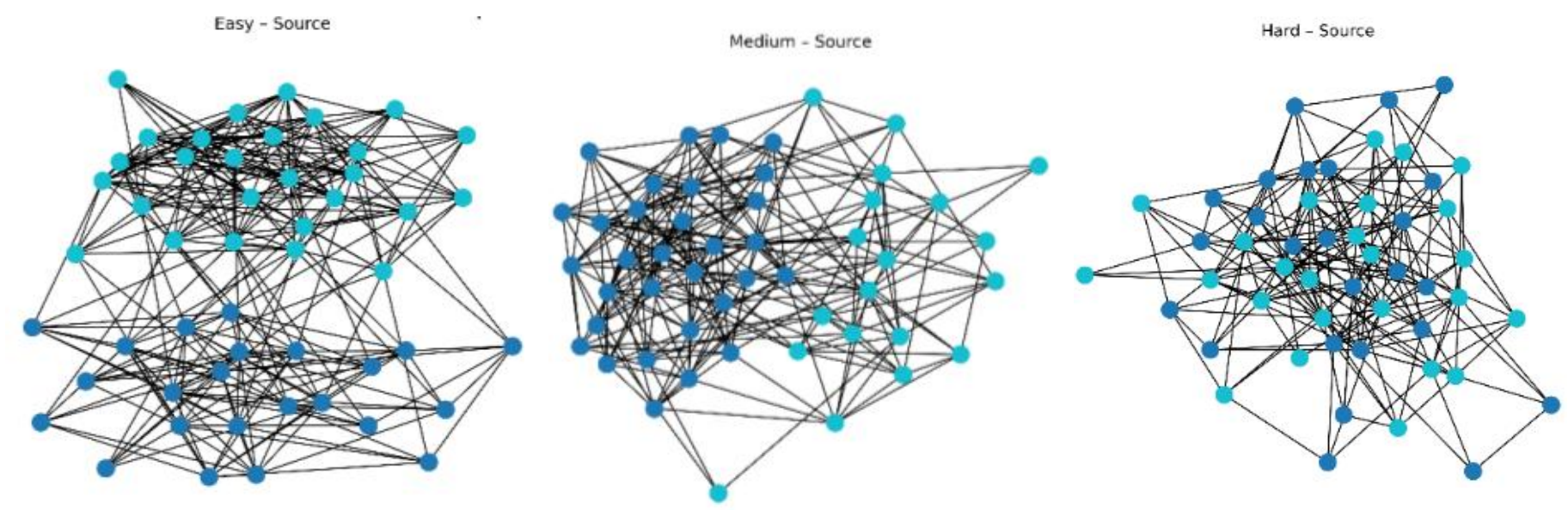

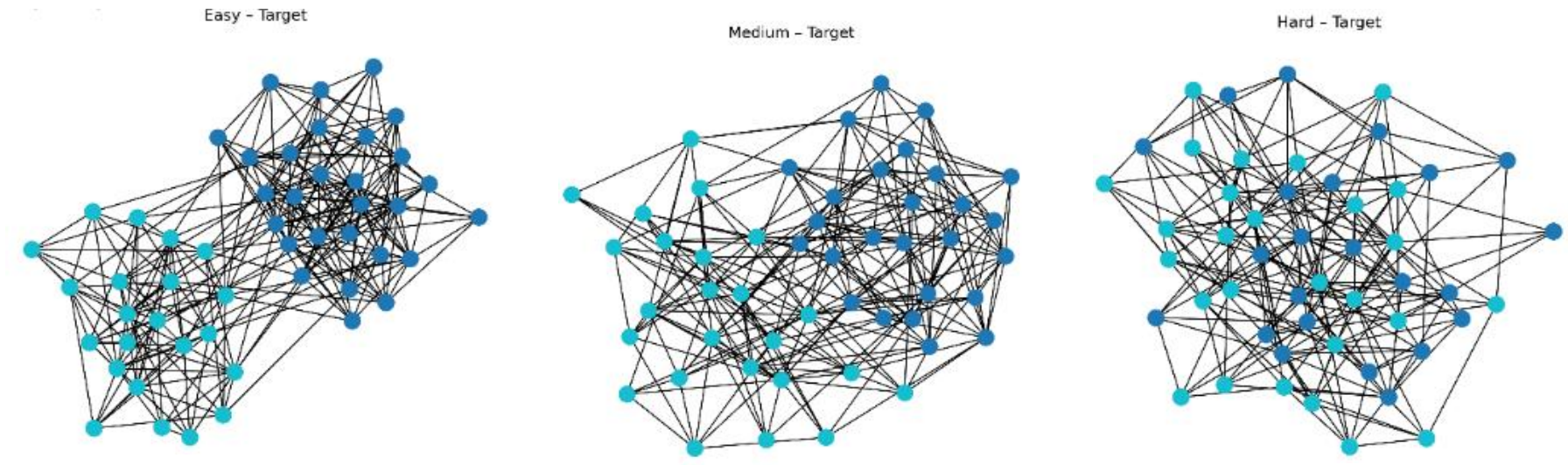

**Fig. 4.** Synthetic benchmark graphs (n=50) at three difficulty levels with node colors indicating cluster labels. Easy, Clear separation. Medium, Moderate mixing. Hard, Sparse, ambiguous structure. Source (left) and target (right) are independently generated.

## 5.2 Comprehensive Performance Analysis

We evaluate transport plan quality using multiple metrics: Accuracy measures label consistency by assigning each target node to the source transporting most mass; ARI [32] is a chance-corrected clustering metric ranging from -1 (worse than random) to 1 (perfect), more sensitive to cluster structure than accuracy; Distance combines feature dissimilarity (weighted by 1-α) and structural discrepancy (weighted by α); Sparsity counts target nodes receiving non-negligible mass, with lower values indicating concentrated transport plans [33].

Tables 1-3 summarize results across difficulties. In easy instances, DsrFGW achieves accuracy gains up to 0.12 and ARI improvements up to 0.243, with strong correlation to distance reductions ($\rho > 0.85$). In medium instances, DsrFGW's advantage becomes pronounced with accuracy gains of 7-20 percentage points and ARI improvements of 0.105-0.239; critically, srFGW often achieves negative ARI while DsrFGW produces positive scores ($\rho = 0.78$). On hard instances, DsrFGW outperforms srFGW in 11 of 12 configurations with accuracy gains of 2-9 percentage points, transforming negative/near-zero ARI values into positive values. Optimal diffusion scales: $\tau \approx 0.5$ for small graphs (n=50, 100) in easy/medium regimes; $\tau \in [0.01, 0.10]$ for n=200; $\tau \in [0.01, 0.20]$ in hard instances to avoid over-smoothing.

**Table 1.** Performance on easy instances (best DsrFGW configuration per α)

| Size | α | srFGW | | DsrFGW | | τ* | Δ Acc | Δ ARI |
|---|---|---|---|---|---|---|---|---|
| | | Acc | ARI | Acc | ARI | | | |
| n=50 | 0.1 | 0.760 | 0.256 | 0.880 | 0.569 | 0.50 | +0.120 | +0.313 |
| | 0.3 | 0.780 | 0.300 | 0.900 | 0.633 | 0.50 | +0.120 | +0.333 |

| Size | α | srFGW Acc | srFGW ARI | DsrFGW Acc | DsrFGW ARI | τ* | Δ Acc | Δ ARI |
|---|---|---|---|---|---|---|---|---|
| | 0.5 | 0.860 | 0.509 | 0.900 | 0.633 | 0.20 | +0.040 | +0.124 |
| | 0.7 | 0.820 | 0.398 | 0.920 | 0.700 | 0.20 | +0.100 | +0.302 |
| n=100 | 0.1 | 0.790 | 0.330 | 0.820 | 0.404 | 0.50 | +0.030 | +0.074 |
| | 0.3 | 0.830 | 0.430 | 0.860 | 0.514 | 0.50 | +0.030 | +0.084 |
| | 0.5 | 0.820 | 0.404 | 0.860 | 0.514 | 0.50 | +0.040 | +0.110 |
| | 0.7 | 0.840 | 0.457 | 0.890 | 0.605 | 0.50 | +0.050 | +0.148 |
| n=200 | 0.1 | 0.755 | 0.256 | 0.760 | 0.267 | 0.50 | +0.005 | +0.011 |
| | 0.3 | 0.745 | 0.236 | 0.775 | 0.299 | 0.01 | +0.030 | +0.063 |
| | 0.5 | 0.765 | 0.277 | 0.785 | 0.322 | 0.01 | +0.020 | +0.045 |
| | 0.7 | 0.770 | 0.288 | 0.770 | 0.288 | 0.01 | 0.000 | 0.000 |

**Table 2.** Performance on medium instances (best DsrFGW configuration per α)

| Size | α | srFGW Acc | srFGW ARI | DsrFGW Acc | DsrFGW ARI | τ* | Acc Δ | ARI Δ |
|---|---|---|---|---|---|---|---|---|
| n=50 | 0.1 | 0.480 | -0.018 | 0.660 | 0.087 | 0.50 | +0.180 | +0.105 |
| | 0.3 | 0.500 | -0.020 | 0.700 | 0.144 | 0.50 | +0.200 | +0.164 |
| | 0.5 | 0.440 | -0.004 | 0.620 | 0.039 | 0.50 | +0.180 | +0.043 |
| | 0.7 | 0.540 | -0.008 | 0.700 | 0.145 | 0.50 | +0.160 | +0.153 |
| n=100 | 0.1 | 0.560 | 0.005 | 0.750 | 0.244 | 0.50 | +0.190 | +0.239 |
| | 0.3 | 0.710 | 0.169 | 0.780 | 0.308 | 0.50 | +0.070 | +0.139 |
| | 0.5 | 0.670 | 0.108 | 0.740 | 0.224 | 0.50 | +0.070 | +0.116 |
| | 0.7 | 0.650 | 0.083 | 0.730 | 0.206 | 0.50 | +0.080 | +0.123 |
| n=200 | 0.1 | 0.570 | 0.015 | 0.710 | 0.172 | 0.05 | +0.140 | +0.157 |
| | 0.3 | 0.600 | 0.035 | 0.705 | 0.164 | 0.05 | +0.105 | +0.129 |
| | 0.5 | 0.570 | 0.015 | 0.725 | 0.199 | 0.05 | +0.155 | +0.184 |
| | 0.7 | 0.605 | 0.039 | 0.680 | 0.125 | 0.10 | +0.075 | +0.086 |

**Table 3.** Performance on hard instances (best DsrFGW configuration per α)

| Size | α | srFGW | | DsrFGW | | τ* | Acc Δ | ARI Δ |
|---|---|---|---|---|---|---|---|---|
| | | Acc | ARI | Acc | ARI | | | |
| n=50 | 0.1 | 0.580 | 0.008 | 0.600 | 0.022 | 0.05 | +0.020 | +0.014 |
| | 0.3 | 0.560 | -0.003 | 0.560 | -0.003 | 0.00 | 0.000 | 0.000 |
| | 0.5 | 0.420 | 0.008 | 0.620 | 0.040 | 0.01 | +0.200 | +0.032 |
| | 0.7 | 0.480 | -0.017 | 0.580 | 0.007 | 0.01 | +0.100 | +0.024 |
| n=100 | 0.1 | 0.500 | -0.010 | 0.590 | 0.023 | 0.20 | +0.090 | +0.033 |
| | 0.3 | 0.540 | -0.004 | 0.570 | 0.010 | 0.20 | +0.030 | +0.014 |
| | 0.5 | 0.580 | 0.016 | 0.580 | 0.016 | 0.10 | 0.000 | 0.000 |
| | 0.7 | 0.550 | 0.000 | 0.570 | 0.010 | 0.20 | +0.020 | +0.010 |
| n=200 | 0.1 | 0.540 | 0.001 | 0.595 | 0.031 | 0.10 | +0.055 | +0.030 |
| | 0.3 | 0.520 | -0.003 | 0.580 | 0.021 | 0.01 | +0.060 | +0.024 |
| | 0.5 | 0.475 | -0.002 | 0.545 | 0.003 | 0.01 | +0.070 | +0.005 |
| | 0.7 | 0.450 | 0.005 | 0.535 | -0.000 | 0.01 | +0.085 | -0.005 |

## 5.3 Cross-Difficulty Comparative Analysis

The robustness of DsrFGW across difficulty levels is summarized in Table 4, which reports relative improvements in both accuracy and ARI over srFGW. A clear pattern emerges: DsrFGW achieves its largest gains in the medium-difficulty regime, with average accuracy improvements typically between 20-30% and ARI improvements often exceeding 100% by transforming negative to positive values. The ARI improvements are particularly striking in medium and hard instances where srFGW produces negative values (worse than random) while DsrFGW achieves positive values (meaningful clustering); for example, in medium difficulty with n=100 and α=0.1, srFGW achieves ARI=0.005 while DsrFGW reaches ARI=0.244, representing a 4780% relative improvement. In hard instances, improvements persist but are more variable due to inherent matching ambiguity under severe noise, with small graphs (n=50) exhibiting the largest relative gains up to 400% for ARI at α=0.5, highlighting the importance of neighborhood aggregation when both signals are weak. For easy instances and larger graphs, relative improvements are modest (often below 20% for accuracy and 30% for ARI), indicating that when the signal-to-noise ratio is already high, standard srFGW is largely sufficient and diffusion offers limited additional benefit.

**Table 4.** DsrFGW relative accuracy and ARI improvement (%) across difficulties

| Size | Easy (Acc/ARI) | Medium (Acc/ARI) | Hard (Acc/ARI) |
|---|---|---|---|
| n=50 | 12.0% / 83.4% | 37.0% / 997.7% | 17.9% / 204.1% |
| n=100 | 4.6% / 25.4% | 16.6% / 1279.5% | 6.8% / 495.0% |
| n=200 | 1.8% / 11.8% | 20.4% / 715.6% | 13.8% / 1112.5% |

## 5.4 Trade-off Analysis

Figures 5-6 illustrate the variation of accuracy, ARI, and distance with respect to α for hard instances and present the Pareto frontiers, where each point represents a specific $(\alpha, \tau)$ configuration. Table 5 shows DsrFGW's advantage strengthens with graph size: for n=50, DsrFGW dominates srFGW in 50% of configurations, increasing to 75% for n=100 and 100% for n=200. The relationship between ARI improvements and geometric consistency shows moderate positive correlation (ρ = 0.52) for hard instances, strengthening for medium (ρ = 0.78) and easy instances (ρ = 0.85), suggesting objectives align more closely when signal quality is higher. In 8 out of 12 hard configurations (66.7%), DsrFGW simultaneously improves all three metrics—accuracy, ARI, and transport distance—representing genuine Pareto improvements rather than trade-offs between objectives.

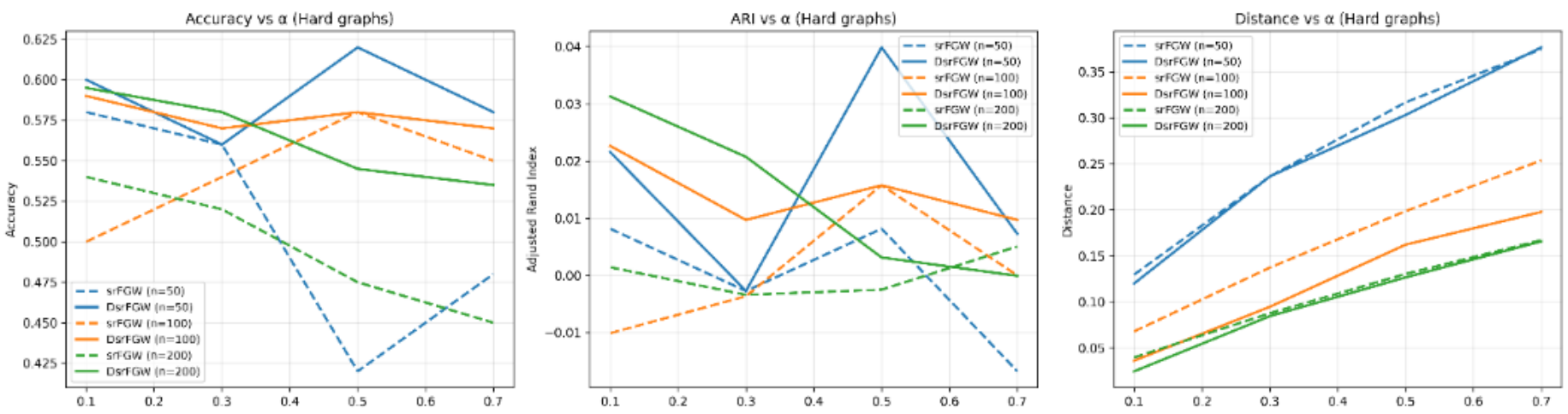

**Fig. 5.** Accuracy, ARI, and distance vs. α for hard instances (n=50, 100, 200). Dashed lines: srFGW, solid lines: DsrFGW with optimal τ.

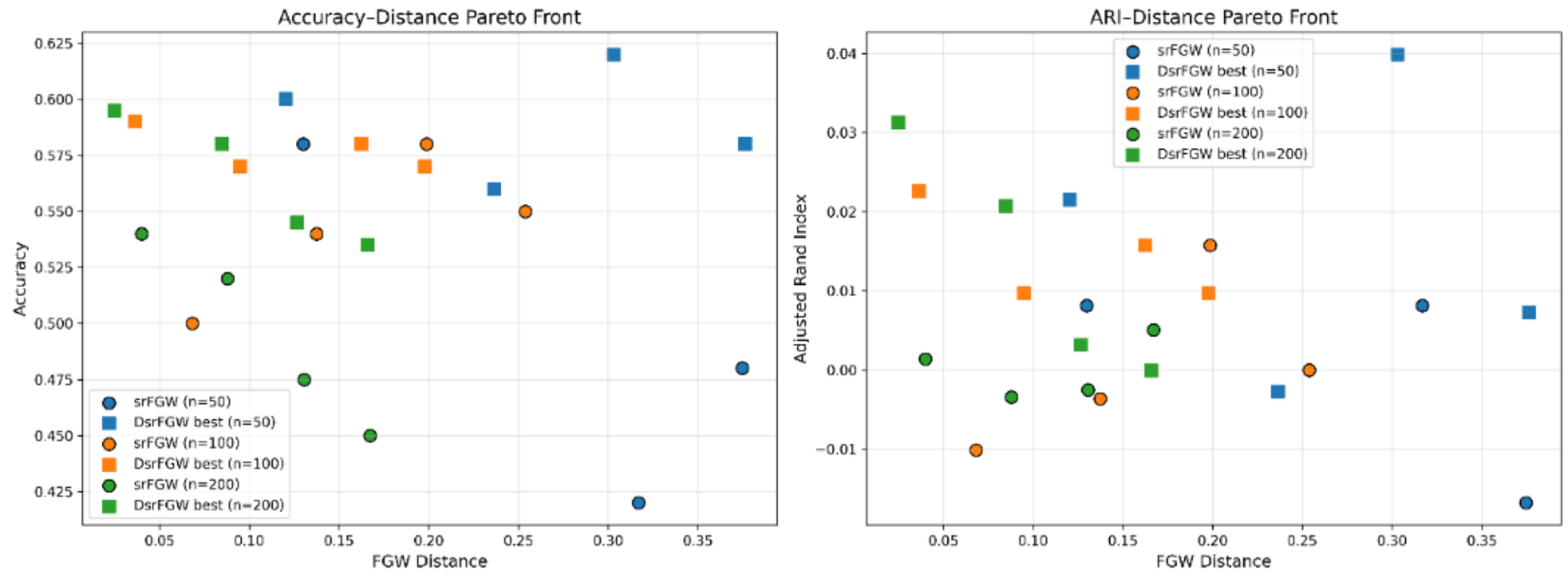

**Fig. 6.** Pareto frontier for hard instances: accuracy-distance (left) and ARI-distance (right). Circles: srFGW, squares: DsrFGW. DsrFGW dominates srFGW, particularly for larger graphs.

**Table 5.** Dominance analysis for hard instances (accuracy-distance space)

| Size | Total Points | DsrFGW Dominates | DsrFGW Equal Acc, Better Dist | srFGW Better |
|---|---|---|---|---|
| n=50 | 4 | 2 (50%) | 1 (25%) | 1 (25%) |
| n=100 | 4 | 3 (75%) | 1 (25%) | 0 (0%) |
| n=200 | 4 | 4 (100%) | 0 (0%) | 0 (0%) |

## 5.5 Sparsity Analysis

Figure 7 illustrates the relationship between accuracy and sparsity for hard instances. In half of examined configurations, DsrFGW achieves higher accuracy and ARI while simultaneously reducing sparsity, indicating diffusion sharpens correspondences rather than merely spreading mass. Most pronounced for n=100 and α=0.5, DsrFGW reduces active transport entries by approximately 22% (from 64 to 50) while maintaining accuracy. The relationship varies with graph size: smaller graphs (n=50) tend toward slightly denser plans (+2 to +4 nodes) while achieving large accuracy gains as diffusion discovers additional meaningful correspondences, while larger graphs (n=100, n=200) often show reduced sparsity with improved performance as diffusion clarifies ambiguous matches. These results demonstrate that DsrFGW concentrates mass on structurally and semantically meaningful correspondences, improving both clustering quality and geometric efficiency.

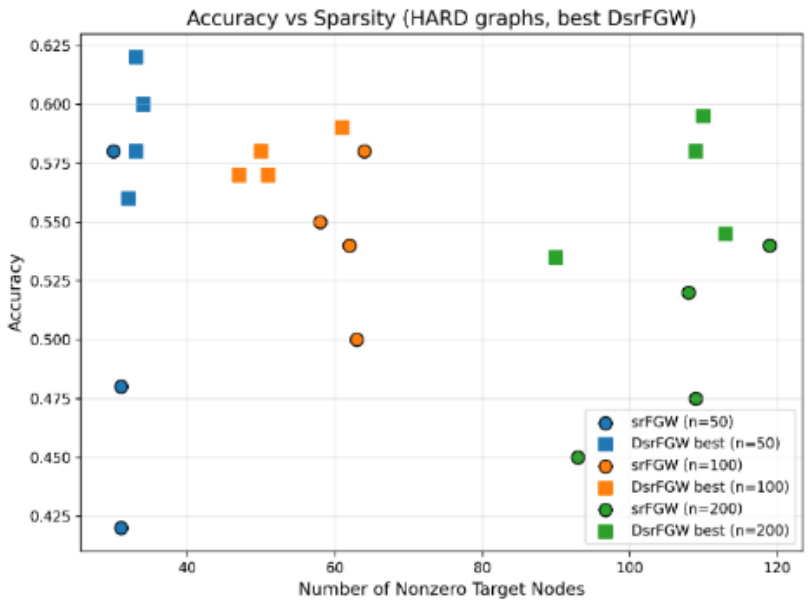

**Fig. 7.** Accuracy vs. sparsity for hard instances. Circles: srFGW, squares: DsrFGW. Colors: blue (n=50), orange (n=100), green (n=200). DsrFGW achieves higher accuracy with size-dependent sparsity increases for small graphs, decreases for large graphs.

## Conclusions, Limitations, and Perspectives

This paper presents DsrFGW, a method for comparing graphs with node attributes by combining graph diffusion, semi-relaxed optimal transport, and fused Gromov-Wasserstein distances. The approach consistently outperforms standard methods across synthetic benchmarks, achieving 7-20 percentage point accuracy improvements in medium-difficulty scenarios and maintaining robust performance even under severe noise. The method shows that incorporating diffusion is crucial (preferred in 97% of configurations), with optimal diffusion scales depending on problem difficulty and larger graphs benefiting more from information aggregation across neighborhoods.

Key limitations include inherited computational complexity from Gromov-Wasserstein (NP-hard quadratic assignment), cubic scaling of diffusion preprocessing, and challenges in parameter selection. Future directions include developing specialized solvers with entropic regularization for GPU acceleration, extending to dynamic graphs for temporal tracking, exploring connections to Bayesian optimization using DsrFGW-based kernels, and leveraging large language models for adaptive parameter selection or learning assignment problems directly through transformers.

**Disclosure of Interests.** The authors declare no conflicts of interest.